\documentclass[conference]{IEEEtran}
\IEEEoverridecommandlockouts

\usepackage{cite}
\usepackage{amsmath,amssymb,amsfonts}
\usepackage{algorithmic}
\usepackage{graphicx}
\usepackage{textcomp}
\usepackage{xcolor}
\usepackage{balance}
\usepackage{tabularx}
\usepackage{graphicx}
\usepackage{adjustbox}
\usepackage{amsmath}
\usepackage{booktabs}  
\usepackage{multirow}
\usepackage{mathrsfs}
\usepackage{enumitem}
\usepackage{stfloats}  
\usepackage{float} 
\usepackage{placeins} 
\usepackage{url}

\begin{document}
\title{MMFformer: Multimodal Fusion Transformer Network for Depression Detection}


\author{Md Rezwanul Haque\textsuperscript{1}, Md. Milon Islam\textsuperscript{1}, S M Taslim Uddin Raju\textsuperscript{1}, Hamdi Altaheri\textsuperscript{1}, \\Lobna Nassar\textsuperscript{2}, and Fakhri Karray\textsuperscript{1,3}

\thanks{\textsuperscript{1}The authors are with the Centre for Pattern Analysis and Machine Intelligence, Department of Electrical and Computer Engineering, University of Waterloo, N2L 3G1, Ontario, Canada. (e-mail: rezwan@uwaterloo.ca{*}, milonislam@uwaterloo.ca, smturaju@uwaterloo.ca, haltaheri@uwaterloo.ca).

\textsuperscript{2}The author is with the School of Engineering and Computing, Department of Computer Science and Engineering, American University of Ras Al Khaimah, Ras Al Khaimah, United Arab Emirates. (e-mail: lobna.nassar@aurak.ac.ae).

\textsuperscript{1,3}The author is with the Centre for Pattern Analysis and Machine Intelligence, Department of Electrical and Computer Engineering, University of Waterloo, N2L 3G1, Ontario, Canada, and Department of Machine Learning, Mohamed bin Zayed University of Artificial Intelligence, Abu Dhabi, United Arab Emirates. (e-mail: karray@uwaterloo.ca, fakhri.karray@mbzuai.ac.ae).}
}
\maketitle

\begingroup
\renewcommand\thefootnote{}

\footnotetext{
\textsuperscript{*}Correspondence to: Md Rezwanul Haque\texttt{<rezwan@uwaterloo.ca>}.
}

\footnotetext{
\textsuperscript{©}\textit{Proceedings of the 2025 IEEE International Conference on Systems, Man, and Cybernetics (SMC), Vienna, Austria. Copyright 2025 by the author(s).}
}
\endgroup

\begin{abstract}

Depression is a serious mental health illness that significantly affects an individual's well-being and quality of life, making early detection crucial for adequate care and treatment. Detecting depression is often difficult, as it is based primarily on subjective evaluations during clinical interviews. Hence, the early diagnosis of depression, thanks to the content of social networks, has become a prominent research area. The extensive and diverse nature of user-generated information poses a significant challenge, limiting the accurate extraction of relevant temporal information and the effective fusion of data across multiple modalities. This paper introduces MMFformer, a multimodal depression detection network designed to retrieve depressive spatio-temporal high-level patterns from multimodal social media information. The transformer network with residual connections captures spatial features from videos, and a transformer encoder is exploited to design important temporal dynamics in audio. Moreover, the fusion architecture fused the extracted features through late and intermediate fusion strategies to find out the most relevant intermodal correlations among them. Finally, the proposed network is assessed on two large-scale depression detection datasets, and the results clearly reveal that it surpasses existing state-of-the-art approaches, improving the F1-Score by 13.92\% for D-Vlog dataset and 7.74\% for LMVD dataset. The code is made available publicly at \url{https://github.com/rezwanh001/Large-Scale-Multimodal-Depression-Detection}.

\end{abstract}


\section{Introduction}

Depression is a major global mental health concern that affects people's psychological well-being and inhibits social development. The World Health Organization (WHO) shows statistics that more than 280 million people worldwide suffer from depression, which was the fourth largest cause of death in 2023 and is expected to become the primary global health burden by 2030 \cite{malhi2018depression}. Due to the complexity of depression and its variability between individuals, early detection is crucial for allowing immediate care and preventing serious health consequences.

In general practice, physicians diagnose depression through interviews utilizing standardized questionnaires. Physicians evaluate patients' feelings through in-person consultations, observe their facial expressions and body language, and listen attentively to their speech and style \cite{bai2025prediction}. As it depends on a physician's own experience and the subjective descriptions of patients regarding their feelings, it can sometimes lack actual objective validity. The verbal moods of a patient may not always align with their emotional state, as physiological indicators such as heart rate and facial expressions are difficult to control and can often provide more insight \cite{huang2023emotion}. Although electroencephalograms (EEGs) and heart rate monitors offer more objective perspectives, they are not always feasible due to the necessity of specific devices and their limited use outside of clinical settings \cite{jiang2024multimodal}. Currently, the growth of social media, especially video blogs (vlogs), has created new possibilities. People often disclose their ideas, emotions, and daily experiences online, exposing feelings that may not appear during clinical evaluations. These videos are rich in facial, vocal, and verbal signals that can accurately convey emotional states more naturally \cite{min2023detecting}.

In recent years, deep learning has been extensively exploited to develop robust frameworks for analyzing complex multimodal data in mental health applications \cite{cao2025deep}. Compared to conventional techniques that depend on manually generated features, current architectures, such as transformers, can automatically recognize complex patterns in spatial (facial expressions) and temporal (speech rhythms) domains. However, these developments pose various problems, since some existing models analyze spatial and temporal information separately, ignoring the dynamic interaction crucial for accurate mental interpretation \cite{qasim2025detection}. Moreover, the fusion of several modalities, including audio and video, presents challenges due to differences in format, timing, and structure \cite{pawlowski2023effective}. To resolve these issues, efficient fusion mechanisms must fuse various sources to preserve their complementary features while retaining critical information.

In this paper, we present a multimodal fusion network, called MMFformer, to detect depression from social media information. To tackle the issues of extracting and fusing spatio-temporal information, we propose a system capable of capturing high-level spatial features from video data utilizing a transformer with residual connections, while synchronously modeling the temporal dynamics of speech signals through a transformer encoder. Moreover, we propose a fusion module that incorporates late and intermediate fusion methods to enhance the relationship between modalities. For empirical experiments, we perform comprehensive tests on depression datasets, D-Vlog and LMVD, where MMFformer demonstrates superior results compared to the state-of-the-art (SOTA) approaches. Our major contributions are summarized as follows.

\begin{enumerate}

  \item A visual feature extraction mechanism utilizing a residual learning transformer architecture is proposed, which allows the extraction of complex spatial patterns from dynamic facial expressions.

  \item An audio processing network employing a transformer encoder is developed to effectively preserve temporal dependencies in speech relevant to depression signals.

  \item A fusion module comprising late transformer fusion, intermediate transformer fusion, and intermediate attention fusion is introduced to improve the interaction between audio and visual modalities.

  \item Comprehensive tests using two publicly accessible datasets, D-Vlog and LMVD, illustrate that our developed system outperforms several current SOTA methods.

\end{enumerate}

The rest of the paper is organized as follows. Section \ref{Related Works} provides an overview of depression detection, focusing on deep learning and transformer-based frameworks. Section \ref{Proposed Architecture for Depression Detection} demonstrates the proposed architecture for depression detection, including video feature extraction, audio feature extraction, and fusion network. Section \ref{Experiments and Analysis} elaborates on the datasets used, along with implementation details, and reports the results of the experiments and their analysis. Finally, Section \ref{Conclusion} makes conclusions and highlights potential future works.

\section{Related Works}
\label{Related Works}

Recent years have shown substantial advances in multimodal depression detection through deep learning approaches applied to vlog data. Current research emphasizes the use of deep learning for depression detection, which is more effective compared to manual feature extraction \cite{tahir2025depression}. Some researchers have used single-modality data for depression recognition, while others have used multimodal data that contain comprehensive information for accurate and reliable depression detection \cite{subuhi2024advancements}. This section briefly outlines relevant research on methodologies applied to depression detection, including deep learning and transformer models using various data modalities.  

\subsection{Deep Learning for Depression Detection}

DepMamba \cite{ye2025depmamba} introduced an audio-visual progressive fusion network based on Mamba to detect depression through multiple data modalities. The architecture combined convolutional neural networks (CNNs) and Mamba to capture local-to-global features across long-range sequences. It incorporated a multimodal collaborative state space model (SSM) to extract both intermodal and intramodal information for each modality. A multimodal enhanced SSM is exploited to further enhance the cohesion between modalities. Experimental results showed that DepMamba achieved an accuracy of 68.87\% on the D-Vlog dataset and 72.13\% on the LMVD dataset. Xing et al. \cite{xing2024emo} presented a multimodal depression detection framework called EMO-Mamba, which employed multimedia data to enhance performance. The technique applied a CNN to extract the spatial attributes of facial features and the local acoustic features from audio. The SSM network is utilized to understand temporal variations and efficiently maintain memory over long time-series. A multimodal fusion framework is proposed to efficiently combine crucial information from various modalities, enhancing overall detection capabilities. In the D-Vlog dataset, EMO-Mamba obtained accuracy, precision, recall, and F1-Score of 75.54\%, 75.79\%, 75.54\%, and 75.66\%, respectively. Shangguan et al. \cite{shangguan2023automatic} proposed a multiple instance learning (MIL) approach for detecting depression using social media data. The proposed MIL architecture is designed to handle long-term sequences of visual data through an attention-based deep long short-term memory (AD-LSTM) network. The AD-LSTM processed fixed-length visual and speech segments to retrieve temporal dimensions of each instance, and the AD-MIL block fused the temporal representations obtained to perform depression detection. Compared with existing benchmarks, the experiments demonstrated that the developed MIL network achieved the highest weighted average precision, recall, and F1-Score of 67.27\%, 67.77\%, and 66.64\%, respectively. In another research,  Zhou et al. \cite{zhou2023caiinet} developed a deep learning based framework called CAIINET for the early detection of depression, utilizing contextual attention and an information interaction mechanism. The proposed system used a contextual attention module with a Bi-LSTM model to capture crucial audio and visual cues at important temporal points. The system incorporated local and global information fusion modules that evaluated the significance and interaction between the extracted attributes at both local and global levels. Experiments on the D-Vlog dataset revealed that CAIINET surpassed current benchmark models, achieving 66.56\%, 66.98\%, and 66.55\% for weighted average precision, recall, and F1-Score, respectively. Kowalewski et al. \cite{kowalewski2023end} compared machine learning and deep learning techniques for depression detection using audio-visual social media content data. The proposed approach applied three learning algorithms, including EfficientNet, neural network, and XGBoost on D-Vlog dataset and obtained the highest F1-Score of 77\% from the XGBoost classifier. 

\subsection{Transformer for Depression Detection}

He et al. \cite{he2025lmtformer} developed a lightweight architecture named LMTformer, aimed to detect the depression from facial videos through a multi-scale transformer. This model retrieved coarse-grained attributes from facial expressions and then processed them through a lightweight multi-scale transformer. The transformer recorded local and global patterns in diverse receptive fields. Moreover, global features are enhanced by a multi-scale global feature fusion approach. Using the LMVD dataset, the proposed network achieved accuracy, precision, recall and F1-Score of 82.76\%, 82.87\%, 82.76\% and 82.74\%, respectively. Further, a video-based depression detection system termed Depressformer is introduced in \cite{he2024depressformer}. The model utilized the video Swin Transformer to enhance the extraction of vital video features. A module focused on depression-specific fine-grained local feature extraction is presented to identify detailed signs of depression. In addition, a depression channel attention fusion block is added to improve the fusion and modeling of the combined features.
The empirical findings demonstrated its performance, obtaining an F1-Score of 0.59 on the D-vlog dataset. Tao et al. \cite{tao2024depmstat} proposed a depression detection model called DepMSTAT, which analyzes audio and visual features from vlog content to identify depression. The spatial-temporal attentional transformer (STAT) block is at the core of the system, designed to capture spatial and temporal relationships within multimodal data effectively. This module extracted spatio-temporal features from individual modalities and then fused them for analyzing vlog-based audio and visual signals. According to experimental results, DepMSTAT achieved precision of 71.53\%, recall of 75.60\%, and F1-Score of 73.51\%. Further, Tao et al. \cite{tao2024depressive} presented a spatio-temporal squeeze transformer (STST) technique to extract relevant semantic features related to depression. The approach employed a transformer encoder to process spatio-temporal data and extract significant features, which are then utilized by a voting-based classifier to detect the depression. The experiments on the D-Vlog dataset achieved an accuracy of 70.70\%, a precision of 72.50\%, a recall of 77.67\%, and an F1-Score of 75\%. Yang et al. \cite{yang2025spike} deployed a computationally efficient hierarchical structure for autonomous depression detection in an Internet of Things (IoT) environment. This framework enabled IoT devices to collaborate in a layered and distributed way to obtain mental health information. The proposed method trained the spike memory transformer (SMT) to capture complex temporal relationships and heterogeneous patterns within data for depression recognition. Experimental results showed that SMT outperformed traditional deep learning methods with an accuracy of 70.73\% for D-Vlog dataset and obtained lower consumption of energy during inference. 

\section{MMFformer Architecture for Depression Detection}

\label{Proposed Architecture for Depression Detection}

\begin{figure*}[!t]

\centerline{\includegraphics[trim={1cm 1.5cm 6cm 1cm}, scale=.66]{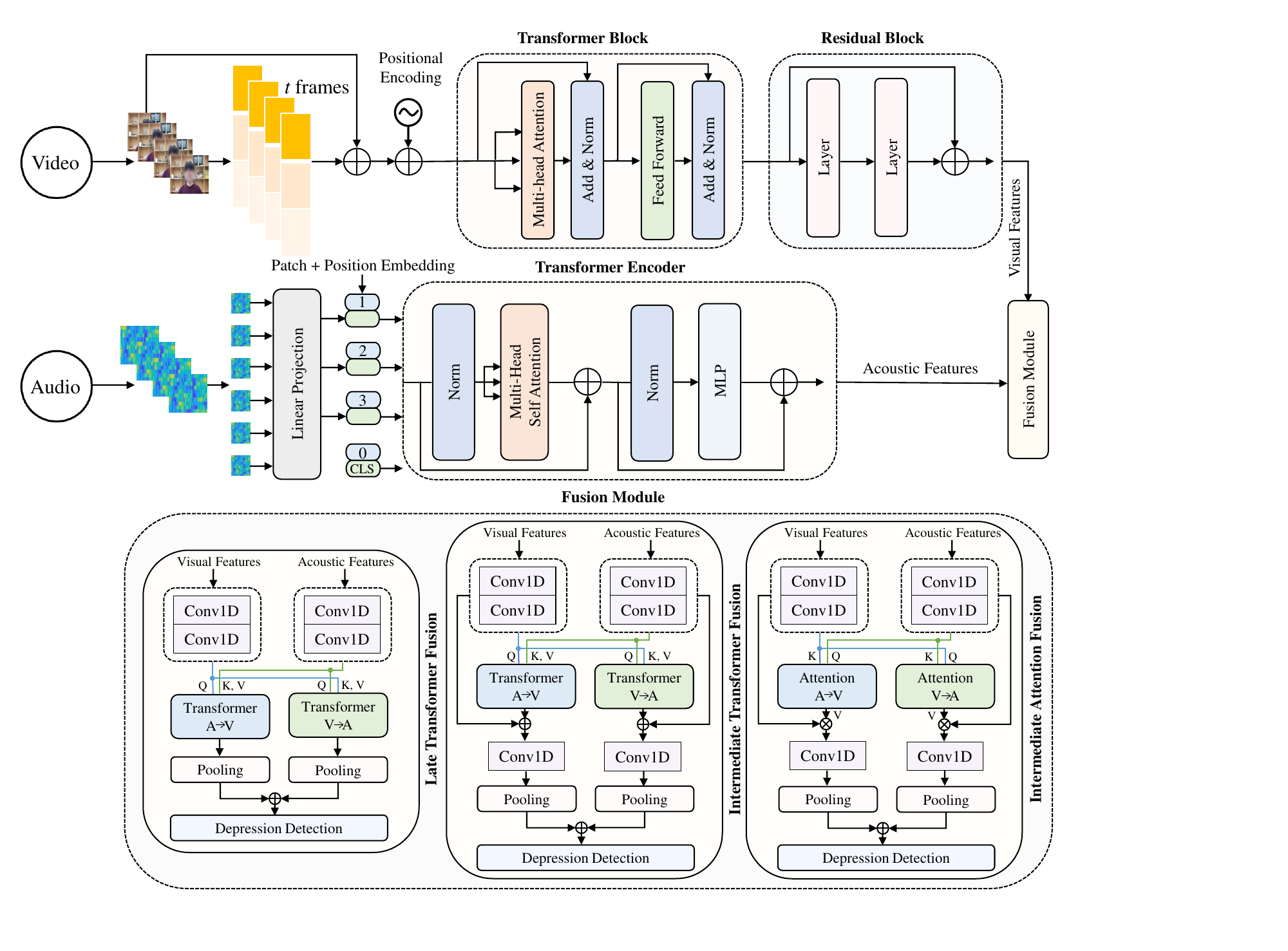}} 

\caption{A brief overview of the MMFformer architecture for multimodal depression detection. The proposed approach consists of multiple modules, including video feature extraction (top part), audio feature extraction (middle part), and late and intermediate fusion (bottom part). Video feature extraction utilizes transformer blocks and residual connections to capture spatial patterns from video clips. Audio feature extraction includes processing audio signals through a transformer encoder to extract meaningful temporal dynamics in speech signals. The fusion module works at late and intermediate stages to capture significant intermodal interactions among the extracted features. Finally, the fused features are fed into a classifier to detect depressive states from the multimodal inputs.}
\label{fig:depression_SA}
\end{figure*}

\subsection{Video Feature Extraction}
In the video feature extraction (as shown in top part of Fig. \ref{fig:depression_SA}), the video data is first pre-processed and then embedded in a high-dimensional space suitable for transformer-based processing. The module utilizes a pre-trained vision transformer (ViT) architecture \cite{chumachenko2024mma} to process the video signals.

Initially, we downsample the video input using a 1D convolution block along the temporal dimension, where $\mathcal{T}$ is the sequence length, and $\mathcal{C}$ is the feature dimension for input tensor $\mathcal{X}_v \in \mathbb{R}^{\mathcal{T} \times \mathcal{C}}$. This operation refines the input resolution to a fixed length $\mathcal{L}$ as mentioned in (\ref{x_down_sampling}).
\begin{equation}
    \widetilde{\mathcal{X}_v} = \mathcal{F}_{d}(\mathcal{X}_v)
    \label{x_down_sampling}
\end{equation}
where $\mathcal{F}_{d}(\cdot)$ represents the sequence of convolution, normalization, and pooling operations. Then, we apply a linear patch embedding ($\mathcal{F}_{emb}$) as in (\ref{li_patch_embd}).
\begin{equation}
    \mathcal{X}_\mathcal{E} = \mathcal{F}_{emb}(\widetilde{\mathcal{X}}_v) = \mathcal{W}_{emb} \widetilde{\mathcal{X}}_v \quad \in \mathbb{R}^{ \mathcal{L} \times \mathcal{D}}
    \label{li_patch_embd}
\end{equation}
where $\mathcal{W}_{emb} \in \mathbb{R}^{\mathcal{C} \times \mathcal{D}}$ and $\mathcal{D}$ are the embedding weight and dimension.

After obtaining the patch embeddings $\mathcal{X}_\mathcal{E}$, a learnable classification token ($\mathcal{T}_{cls}$) is added at the beginning of the embedding sequence as shown in (\ref{p_cancat}).
\begin{equation}
        \mathcal{X}_v^{(0)} = \mathcal{T}_{cls} \oplus \mathcal{X}_\mathcal{E} \quad \in \mathbb{R}^{\mathcal{(L+{\text{1}})} \times \mathcal{D}}
    \label{p_cancat}
\end{equation}
In addition, learnable positional encoding $\mathcal{P} \in \mathbb{R}^{1 \times (\mathcal{L}+1) \times \mathcal{D}}$ is incorporated to capture spatial information as in (\ref{add_pos_emb}).
\begin{equation}
    \mathcal{X}_v^{(1)} = \mathcal{X}_v^{(0)} + \mathcal{P}
    \label{add_pos_emb}
\end{equation}

The token-augmented sequence $\mathcal{X}_v^{(1)}$ is subsequently passed through $\mathcal{N}$ transformer blocks ($n = 1, \ldots, \mathcal{N}$). In each block, self-attention ($\mathcal{Z}^{n}$) is computed as described in (\ref{trans_blocks}), where the queries, keys, and values are represented as $\mathcal{Q}^{n} = \mathcal{X}_v^{(n)} \mathcal{W}^\mathcal{Q}$, $\mathcal{K}^{n} = \mathcal{X}_v^{(n)} \mathcal{W}^\mathcal{K}$, and $\mathcal{V}^{n} = \mathcal{X}_v^{(n)} \mathcal{W}^\mathcal{V}$, respectively.

\begin{align}
    \mathcal{Z}^{n} &= \mathcal{F}_{soft}\left(\frac{\mathcal{Q}^{n}(\mathcal{K}^{n})^\top}{\sqrt{d}}\right)\mathcal{V}^{n}
    \label{trans_blocks}
\end{align}
here $\mathcal{W}^\mathcal{Q},\, \mathcal{W}^\mathcal{K},\, \mathcal{W}^\mathcal{V} \in \mathbb{R}^{\mathcal{D} \times \mathcal{D}}$ are the learnable weight matrices, $\mathcal{F}_{soft}$ is the softmax activation, and $d$ is the dimension for each head. 

A residual connection followed by a multi-layer perceptron (MLP) with layer normalization is applied to the output of each transformer block as shown in (\ref{res_blocks}).
\begin{equation}
    \mathcal{X}_v^{(n+1)} = \mathcal{F}_{mlp}(\mathcal{Z}^{n} + \mathcal{X}_v^{(n)}) + \left( \mathcal{Z}^{n} + \mathcal{X}_v^{(n)} \right)
    \label{res_blocks}
\end{equation}
This mechanism is applied repeatedly, where the output of the final transformer block, $\mathcal{X}_v^{(o)} \in \mathbb{R}^{(\mathcal{L}+1) \times \mathcal{D}} $ serves as the high-level visual feature representation from the video input.

\subsection{Audio Feature Extraction}
The module processes an input audio waveform \( \mathcal{X}_{a} \in \mathbb{R}^{\mathcal{S}} \) ($\mathcal{S}$ denotes the number of samples) through a series of transformations, including linear projection, patch and positional embedding, and transformer encoding (as illustrated in middle part of Fig. \ref{fig:depression_SA}). 

Initially, the waveform is transformed into a time-frequency representation \( \mathcal{X}_{f} \in \mathbb{R}^{\mathcal{F} \times \mathcal{T}} \), where \( \mathcal{F} \) and \( \mathcal{T} \) represent the number of frequency bins and time frames. To ensure consistent input dimensions, \( \mathcal{X}_{f} \) is projected to a fixed-size matrix \( \mathcal{X}'_{f} \in \mathbb{R}^{\mathcal{F}' \times \mathcal{T}'} \) via a learnable linear projection function \(\mathcal{F}_{lnp}\) as in (\ref{xa_down_sampling}).
\begin{equation}
    \mathcal{X}'_{f} = \mathcal{F}_{lnp}(\mathcal{X}_{f}; \mathcal{F}', \mathcal{T}', \theta_d)
    \label{xa_down_sampling}
\end{equation}
where \(\theta_d\) represents the parameters of convolutional and pooling operations. In particular, a 1D convolution adjusts the frequency dimension, followed by batch normalization and adaptive average pooling to standardize the time dimension to \(\mathcal{T}'\). The resulting matrix is then reshaped into a single-channel 2D input \( \mathcal{X}'_{f} \in \mathbb{R}^{1 \times \mathcal{F}' \times \mathcal{T}'} \).

The obtained feature matrix \( \mathcal{X}'_{f} \) is partitioned into overlapping patches using a 2D convolutional layer to generate patch embeddings \( \mathcal{X}_{p} \in \mathbb{R}^{\mathcal{M} \times \mathcal{D}} \). The convolution operates with patch sizes \((p_f, p_t)\) and strides \((s_f, s_t)\), generating a feature map with spatial dimensions \( h = \lfloor (\mathcal{F'} - p_f)/s_f \rfloor + 1 \) and \( w = \lfloor (\mathcal{T'} - p_t)/s_t \rfloor + 1 \), resulting in \(\mathcal{M} = h \times w\) patches. The resulting feature map in 3D tensor form is flattened and transposed using the operation \( \mathcal{F}_{flt}(\cdot) \) as in (\ref{feat_P}), where \( \tilde{\mathcal{X}}_{p} \) denotes the intermediate output of the convolution that makes the sequence appropriate for transformer encoder.
\begin{equation}
    \mathcal{X}_{p} = \mathcal{F}_{flt}(\tilde{\mathcal{X}}_{p}) \in \mathbb{R}^{\mathcal{M} \times \mathcal{D}}
    \label{feat_P}
\end{equation}

A base positional embedding matrix \( \mathcal{X}_{e_{{base}}} \in \mathbb{R}^{\mathcal{D} \times h_{{base}} \times w_{{base}}} \), following the audio spectrogram transformer (AST) \cite{gong2021ast}, is resized using bilinear interpolation to align with the patch grid size \((h, w)\) as mentioned in (\ref{flat}).
\begin{equation}
    \mathcal{X}_e = \mathcal{F}_{inp}(\mathcal{X}_{e_{{base}}}; (h, w)) \in \mathbb{R}^{\mathcal{M} \times \mathcal{D}}
    \label{flat}
\end{equation}
Moreover, two special tokens are added to the sequence: a classification token \( x_{{cls}} \) and a distillation token \( x_{{dist}} \), each with fixed positional embeddings \( e_{{cls}} \) and \( e_{{dist}} \). The final embedded sequence is generated as in (\ref{token_eq}).
\begin{equation}
\begin{split}
\mathcal{X}_{pe} = [x_{{cls}} + e_{{cls}},\; x_{{dist}} + e_{{dist}}, \\
\mathcal{X}_{p_1} + \mathcal{X}_{e_1},\; \dots,\; \mathcal{X}_{p_{\mathcal{M}}} + \mathcal{X}_{e_{\mathcal{M}}}] \in \mathbb{R}^{(\mathcal{M} + 2) \times \mathcal{D}}
\end{split}
\label{token_eq}
\end{equation}

The embedded sequence \( \mathcal{X}_{pe} \), obtained from (\ref{token_eq}), is processed by a transformer encoder comprising several identical layers. Each layer consists of a multi-head self-attention (\(\mathcal{F}_{mhsa}\)) followed by a feed-forward network. Each sub-layer is preceded by layer normalization (\(\mathcal{F}_{ln}\)) and followed by a residual connection. The computations within a single transformer layer are shown in (\ref{transformer_layer}).
\begin{equation}
\begin{split}
    \mathcal{U} &= \mathcal{F}_{ln}\left( \mathcal{X}_{pe} + \mathcal{F}_{mhsa}(\mathcal{X}_{pe}) \right), \\
    \mathcal{Z} &= \mathcal{F}_{ln}\left( \mathcal{U} + \mathcal{F}_{mlp}(\mathcal{U}) \right)
\end{split}
\label{transformer_layer}
\end{equation}
where \( \mathcal{Z} \) corresponds to the final acoustic output sequence \( \mathcal{X}_{a}^{(o)} \in \mathbb{R}^{(\mathcal{M} + 2) \times \mathcal{D}} \), capturing temporal feature representations.

\subsection{Fusion Module}
In this section, the proposed fusion module is described as illustrated in fusion module of bottom part of Fig.~\ref{fig:depression_SA}.

\subsubsection{Late Transformer Fusion}
This architecture fuses the extracted visual and acoustic features using transformer blocks. Each network employs its own transformer block to perform cross-modal fusion. The video network takes visual features \( \mathcal{X}_v^{(o)} \), while the audio network processes acoustic features \( \mathcal{X}_a^{(o)} \), and fusion happens by combining features from one modality into the other. Each network processes its respective features through a series of Conv1D layers. The outputs of each transformer block are then pooled, concatenated and fed into a final depression detection layer. 

For the acoustic network, the transformer block takes the visual network representation \( \mathcal{X}_v^{(o)} \) as input to compute the keys and values, while queries are derived from the features of the acoustic network \( \mathcal{X}_a^{(o)} \). The self-attention mechanism is computed in (\ref{late_eq}).
\begin{equation}
\mathcal{O} = \mathcal{F}_{soft}\left( \frac{ \mathcal{X}_a^{(o)} \mathcal{W}_q \mathcal{W}_k^\top (\mathcal{X}_v^{(o)})^\top }{ \sqrt{d} } \right) \mathcal{X}_v^{(o)} \mathcal{W}_v
\label{late_eq}
\end{equation}
where \( \mathcal{W}_q \), \( \mathcal{W}_k \), and \( \mathcal{W}_v \) are the weight matrices for queries, keys, and values, respectively. This operation fuses visual information into the acoustic network (\( V \to A \)). Similarly, the visual network transformer block computes queries from \( \mathcal{X}_v^{(o)} \), and keys and values from \( \mathcal{X}_a^{(o)} \), allowing fusion of acoustic information into the visual network (\( A \to V \)).

\subsubsection{Intermediate Transformer Fusion}
This module introduces cross-modal fusion in an intermediate stage, allowing an earlier interaction between visual and acoustic features. The extracted features are input into two Conv1D layers in each branch to obtain the intermediate representations \( \mathcal{X}_v^{(o)} \) and \( \mathcal{X}_a^{(o)} \), which are passed through separate transformer blocks for cross-modal fusion as shown in (\ref{late_eq}).

In the acoustic network, queries are computed from \( \mathcal{X}_a^{(o)} \), while keys and values are derived from \( \mathcal{X}_v^{(o)} \), enabling visual-to-acoustic information transfer. However, the visual network receives acoustic features through a symmetric operation. The outputs of each network are then fused separately and passed through an additional Conv1D layer. The refined features then pass through the pooling, concatenation, and final classification layer for depression detection. 

\subsubsection{Intermediate Attention Fusion}
This architecture presents an attention-based fusion at an intermediate level, enabling cross-modal interaction without directly fusing feature representations. The extracted features are fed into two Conv1D layers separately, resulting in visual and acoustic features \( \mathcal{X}_v^{(o)} \) and \( \mathcal{X}_a^{(o)} \), which are processed through attention mechanisms by exploiting dot-product similarity to highlight mutually relevant features.

In the acoustic network, queries are computed from \( \mathcal{X}_a^{(o)} \) and keys from \( \mathcal{X}_v^{(o)} \). The scaled dot-product attention is calculated in (\ref{inter_attn}).
\begin{equation}
\mathcal{O} = \mathcal{F}_{soft}\left(\frac{\mathcal{X}_a^{(o)} \mathcal{W}_q \mathcal{W}_k^\top (\mathcal{X}_v^{(o)})^\top}{\sqrt{d}}\right)
\label{inter_attn}
\end{equation}
The softmax operation emphasizes the most salient features of the visual modality relative to the acoustic features (\( V \to A \)). The visual network performs the same mechanism to compute attention from acoustic features (\( A \to V \)).

The attention vector for the visual network is calculated as \( v_v = \sum_{i=1}^{N_v} \mathcal{O}[:, i] \), capturing the most relevant visual attributes based on their alignment with acoustic features. The acoustic network follows a symmetric process. The attention-weighted outputs are refined through an additional Conv1D layer, pooled, concatenated, and passed to the depression detection layer.

\section{Experiments and Analysis}

\label{Experiments and Analysis}

This section presents an extensive set of experiments to assess the performance of the proposed network for detecting depression. We begin by exploring diverse combinations of features derived using video and audio networks and different methods for fusing these features. From these preliminary results, the most promising designs are selected and their performance is evaluated against several existing models. Ablation studies are performed to gain insight into the distinct effects of various fusion methods concerning the proposed architecture. Lastly, cross-corpus experiments are conducted to evaluate generalizability across multiple datasets.  

\subsection{Datasets}

\subsubsection{D-Vlog}

This research used the D-Vlog dataset \cite{yoon2022d}, a publicly accessible repository of YouTube vlogs collected for depression detection. The dataset includes 961 vlogs, approximately 160 hours of video, recorded by 816 distinct individuals, containing both depressive and normal data. The videos were collected using specific keywords such as “depression vlog” or “daily vlog” and then manually annotated to assess whether the speaker has symptoms of current depression. The dataset provides acoustic features obtained through OpenSMILE utilizing the extended Geneva Minimalistic Acoustic Parameter Set (eGeMAPS) and visual information such as face landmarks retrieved via Dlib. These features are sampled on a per-second interval, making them appropriate for temporal analysis. D-Vlog is unique for collecting real-world, unscripted videos where individuals openly share their daily lives and mental health.

\begin{table*}[b]
\caption{Performances of MMFformer for Depression Detection from Multimodal Vlog Data. Each Model is Run Ten Times to Obtain the Results (Mean $\pm$ Std). Bold Represents the Best and Underline Indicates the Second-best.}
\label{tab-proposed_models}
\centering
\begin{adjustbox}{width=\textwidth}
\begin{tabular}{c c c c c c c | c c c c}
\toprule
\textbf{Dataset} & \textbf{Modalities} & \textbf{Fusion} & \textbf{WAA} & \textbf{WAP} & \textbf{WAR} & \textbf{WAF1} & \textbf{UAA} & \textbf{UAP} & \textbf{UAR} & \textbf{UAF1} \\
\midrule

 & A & -- & 0.7814 $\pm$ 0.039 & 0.8786 $\pm$ 0.030 & 0.9165 $\pm$ 0.031 & 0.8955 $\pm$ 0.018 & 0.7638 $\pm$ 0.044 & 0.8969 $\pm$ 0.028 & 0.9297 $\pm$ 0.028 & 0.9115 $\pm$ 0.018 \\

 & V & -- & 0.7214 $\pm$ 0.051 & 0.8438 $\pm$ 0.031 & 0.9041 $\pm$ 0.037 & 0.8704 $\pm$ 0.021 & 0.6912 $\pm$ 0.064 & 0.8682 $\pm$ 0.028 & 0.9188 $\pm$ 0.034 & 0.8904 $\pm$ 0.021 \\

 \cline{2-11}

\multirow{1}{*}{D-Vlog} & \multirow{3}{*} {A+V} & LT & 0.7958 $\pm$ 0.031 & 0.8779 $\pm$ 0.027 & \underline{0.9367 $\pm$ 0.020} & 0.9046 $\pm$ 0.012 & 0.7731 $\pm$ 0.042 & 0.8966 $\pm$ 0.026 & \textbf{0.9473 $\pm$ 0.016} & 0.9196 $\pm$ 0.011 \\

 & & IT & \textbf{0.8108 $\pm$ 0.026} & \textbf{0.8924 $\pm$ 0.024} & 0.9308 $\pm$ 0.037 & \textbf{0.9092 $\pm$ 0.014} & \textbf{0.7957 $\pm$ 0.029} & \textbf{0.9088 $\pm$ 0.024} & 0.9432 $\pm$ 0.028 & \textbf{0.9239 $\pm$ 0.009} \\

 &  & IA & \underline {0.8030 $\pm$ 0.029} & \underline {0.8806 $\pm$ 0.019} & \textbf{ 0.9380 $\pm$ 0.026} & \underline{0.9071 $\pm$ 0.014} & \underline{0.7776 $\pm$ 0.037} & \underline{0.8995 $\pm$ 0.017} & \underline{0.9471 $\pm$ 0.024} & \underline{0.9215 $\pm$ 0.014} \\

\midrule

 & A & -- & 0.7175 $\pm$ 0.024 & 0.8621 $\pm$ 0.023 & 0.8563 $\pm$ 0.030 & 0.8575 $\pm$ 0.011 & 0.7143 $\pm$ 0.024 & 0.8572 $\pm$ 0.024 & 0.8502 $\pm$ 0.036 & 0.8519 $\pm$ 0.018 \\

 & V & -- & 0.6905 $\pm$ 0.055 & 0.8452 $\pm$ 0.022 & 0.8515 $\pm$ 0.074 & 0.8437 $\pm$ 0.043 & 0.6914 $\pm$ 0.053 & 0.8399 $\pm$ 0.023 & 0.8478 $\pm$ 0.070 & 0.8393 $\pm$ 0.039 \\

 \cline{2-11}

\multirow{1}{*}{LMVD}  & \multirow{3}{*} {A+V} & LT & \textbf{0.8071 $\pm$ 0.022} & \textbf{0.9013 $\pm$ 0.018} & \textbf{0.9112 $\pm$ 0.034} & \textbf{0.9048 $\pm$ 0.013} & \textbf{0.8089 $\pm$ 0.022} & \textbf{0.8966 $\pm$ 0.025} & \textbf{0.9090 $\pm$ 0.031} & \textbf{0.9014 $\pm$ 0.016} \\

 &  & IT & \underline {0.8035 $\pm$ 0.029} & 0.8994 $\pm$ 0.018 & \underline{0.9072 $\pm$ 0.028} & \underline{0.9024 $\pm$ 0.016} & 0.8019 $\pm$ 0.031 & 0.8955 $\pm$ 0.022 & 0.9031 $\pm$ 0.033 & \underline{0.8984 $\pm$ 0.021} \\

 &  & IA & 0.8023 $\pm$ 0.022 & \underline{0.9006 $\pm$ 0.020} & 0.9064 $\pm$ 0.034 & 0.9019 $\pm$ 0.013 & \underline{0.8032 $\pm$ 0.022} & \underline{0.8960 $\pm$ 0.025} & \underline{0.9036 $\pm$ 0.033} & 0.8983 $\pm$ 0.016 \\

\bottomrule
\end{tabular}
\end{adjustbox}
\end{table*}

\begin{table}[t]
\caption{Comparison with the SOTA Methods on D-Vlog and LMVD Datasets for Depression Detection.}
\begin{center}
\resizebox{\columnwidth}{!}{%
\begin{tabular}{c c c c c c}
\hline
\textbf{Methods} & \textbf{Datasets} & \textbf{Acc} & \textbf{Pr} & \textbf{Rc} & \textbf{F1} \\
\hline
Ye et al. \cite{ye2025depmamba} & \multirow{2}{*} {LMVD} & 0.7213 & 0.7018 & 0.7656 & 0.7320 \\
He et al. \cite{he2025lmtformer} &  & \textbf{0.8276} & \underline{0.8287} & \underline{0.8276} & \underline{0.8274} \\
\hline
Ye et al. \cite{ye2025depmamba} & \multirow{9}{*} {D-Vlog} & 0.6887 & 0.6819 & \underline{0.8699} & 0.7644 \\
Xing et al. \cite{xing2024emo} &  & \underline{0.7554} & \underline{0.7579} & 0.7554 & 0.7566 \\
Shangguan et al. \cite{shangguan2023automatic} &  & - & 0.6727 & 0.6777 & 0.6664 \\
Zhou et al. \cite{zhou2023caiinet} &  & - & 0.6656 & 0.6698 & 0.6655 \\
Kowalewski et al. \cite{kowalewski2023end} &  & - & 0.7100 & 0.8400 & \underline{0.7700} \\
He et al. \cite{he2024depressformer} &  & 0.6500 & 0.6400 & 0.5400 & 0.5900 \\
Tao et al. \cite{tao2024depmstat} &  & - & 0.7153 & 0.7560 & 0.7351 \\
Tao et al. \cite{tao2024depressive} &  & 0.7070 & 0.7250 & 0.7767 & 0.7500 \\
Yang et al. \cite{yang2025spike} &  & 0.7073 & - & - & - \\
\hline
\multirow{2}{*} {\textbf{MMFformer}} & D-Vlog & \textbf{0.8108} & \textbf{0.8924} & \textbf{0.9380} & \textbf{0.9092} \\
 & LMVD & \underline {0.8071} & \textbf{0.9013} & \textbf{0.9112} & \textbf{0.9048} \\
\hline
\end{tabular}%
}
\label{tab-sota_res_comp}
\end{center}
\end{table}

\subsubsection{LMVD}

The large-scale multimodal vlog dataset (LMVD) \cite{he2024lmvd} is used to evaluate the performance of MMFformer, a newly collected dataset for detecting depression in everyday contexts. The dataset comprises 1,823 vlog samples, around 214 hours of video, collected from 1,475 people across four platforms, including Bilibili, TikTok, Sina Weibo, and YouTube. Each video in the dataset is classified as either depressed or non-depressed, following a manual assessment by volunteers and validation by clinical professionals. The dataset provides comprehensive multimodal features for analysis, including audio embeddings derived from VGGish and visual features such as facial action units, face landmarks, eye gaze, and head pose. LMVD is a heterogeneous dataset of real-world significance, as the videos collected are spontaneous and self-recorded, accurately depicting real-user behavior rather than being generated in laboratory-controlled environments.


\subsection{Implementation Details and Evaluation Metrics}

The MMFformer is developed and trained using the PyTorch deep learning framework in Python. The performance of the developed system is validated through a 10-fold cross-validation. The batch size is set to 16 during training, and the maximum number of training epochs is 225. For optimization, we choose Adam as the optimizer with a learning rate = 1e-5, weight decay = 0.1, and epsilon = 1e-8. We also implemented an early stopping mechanism at 15 epochs that stops training when validation performance stops improving to prevent overfitting. Our proposed architecture is evaluated on two 48 GB NVIDIA RTX A6000 GPUs.

Four widely recognized performance metrics such as accuracy (Acc), precision (Pr), recall (Rc), and F1-Score (F1) are used to assess the performance. These metrics are able to evaluate overall performance in both balanced and unbalanced datasets. To better understand how well the model performs across different class distributions, we report both weighted average (WA) and unweighted average (UA) for each metric. 

\subsection{Depression Detection Results}

Table~\ref{tab-proposed_models} presents the performance of MMFformer for depression detection on the D-Vlog and LMVD datasets. The results are reported using video (V), audio (A), and audio+video (A+V) modalities with three fusion strategies: late transformer (LT), intermediate transformer (IT), and intermediate attention (IA). Each experiment was carried out ten times, with the mean and standard deviation (mean $\pm$ std) reported for all performance metrics.

For both datasets, it is found that multimodal fusion outperforms unimodal approaches. On the D-Vlog dataset, the IT fusion architecture achieves the highest WAA of 0.8108, WAP of 0.8924, and WAF1 of 0.9092, while the IA fusion receives the highest WAR of 0.9380. Similarly, IT network obtains UAA of 0.7957, UAP of 0.9088, and UAF1 of 0.9239 for unweighted metrics, while IA fusion performs the best in UAR of 0.9471. On the LMVD dataset, the LT model scores the highest performance across all weighted and unweighted metrics, including WAA of 0.8071, WAP of 0.9013, WAR of 0.9112, and WAF1 of 0.9048. The IT and IA models perform closely, with IA slightly outperforming IT in WAP and UAA.

Overall, the results highlight the enhanced performance of fusion-based multimodal learning methods over unimodal baselines. Among fusion strategies, IT has the best performance with the D-Vlog dataset, while LT has the highest performance with the LMVD dataset. The findings indicate that dataset features affect the optimal fusion method, highlighting the significance of customized architecture for multimodal depression detection.

\subsection{Comparison Results}

To evaluate the efficiency of our proposed architecture, we compared its performance with several SOTA approaches in the D-Vlog and LMVD datasets (as summarized in Table~\ref{tab-sota_res_comp}). On the LMVD dataset, our model achieves an F1-Score of 0.9048 and a precision of 0.9013, outperforming all existing methods, including the work presented in \cite{he2025lmtformer}. That work records the second-best F1-Score of 0.8274 and precision of 0.8287, marking a relative improvement of 7.74\% in F1-Score and 7.26\% in precision. Although the system developed in \cite{he2025lmtformer} achieved the highest accuracy of 0.8276, our approach demonstrates superior precision of 0.9013 and recall of 0.9112, indicating better reliability for depression detection. On the D-Vlog dataset, MMFformer outperforms existing methods across all evaluation metrics. An accuracy of 0.8108, a precision of 0.8924, a recall of 0.9380, and an F1-Score of 0.9092 are obtained; all of which surpass previously reported results. The proposed system achieves a relative increase of 5.54\% in accuracy, 13.92\% in F1-Score, and 6.81\% in recall compared to prior methods. The system developed in \cite{tao2024depressive} achieved an F1-Score of 0.7500, with noticeably lower accuracy and precision. While some methods, such as \cite{ye2025depmamba} and \cite{kowalewski2023end} reported competitive recall and F1-Scores, their precision values were comparatively low.


\subsection{Ablation Study}
\begin{table*}[b]
\caption{Results of Ablation Studies for MMFformer on the Video and Audio Data to Detect Depression.}
\label{tab-ablation_study}
\centering
\begin{adjustbox}{width=\textwidth}
\begin{tabular}{c c c c c c c | c c c c}
\toprule
\textbf{Dataset} & \textbf{Modalities} & \textbf{Fusion} & \textbf{WAA} & \textbf{WAP} & \textbf{WAR} & \textbf{WAF1} & \textbf{UAA} & \textbf{UAP} & \textbf{UAR} & \textbf{UAF1} \\
\midrule

\multirow{4}{*}{D-Vlog} & \multirow{4}{*} {A+V} & Add    & \underline{0.7606 $\pm$ 0.032} & \underline{0.8658 $\pm$ 0.024} & \textbf{0.9112 $\pm$ 0.038} & \textbf{0.8860 $\pm$ 0.015} & \underline{0.7385 $\pm$ 0.038} & \underline{0.8866 $\pm$ 0.023} & \textbf{0.9255 $\pm$ 0.033} & \textbf{0.9039 $\pm$ 0.015} \\
                        &  & Multi  & 0.7045 $\pm$ 0.059 & 0.8032 $\pm$ 0.124 & 0.8813 $\pm$ 0.091 & 0.8333 $\pm$ 0.099 & 0.6605 $\pm$ 0.095  & 0.8123 $\pm$ 0.163  & 0.8756 $\pm$ 0.136  & 0.8365 $\pm$ 0.146  \\
                        &  & Concat & \textbf{0.7652 $\pm$ 0.033} & \textbf{0.8794 $\pm$ 0.017} & \underline{0.8890 $\pm$ 0.035} & \underline{0.8833 $\pm$ 0.019} & \textbf{0.7512 $\pm$ 0.035} & \textbf{0.8985 $\pm$ 0.015} & \underline{0.9063 $\pm$ 0.032} & \underline{0.9016 $\pm$ 0.018} \\
                        &  & TF     & 0.6734 $\pm$ 0.095 & 0.6796 $\pm$ 0.241 & 0.8241 $\pm$ 0.175 & 0.7357 $\pm$ 0.218 & 0.6189 $\pm$ 0.115 & 0.6893 $\pm$ 0.270 & 0.8132 $\pm$ 0.207 & 0.7371 $\pm$ 0.248 \\

\midrule

\multirow{4}{*}{LMVD} & \multirow{4}{*} {A+V}& Add  & \textbf{0.7930 $\pm$ 0.029} & \underline{0.8891 $\pm$ 0.017} & \textbf{0.9120 $\pm$ 0.021} & \textbf{0.8998 $\pm$ 0.014} & \textbf{0.7937 $\pm$ 0.029} & \underline{0.8846 $\pm$ 0.021} & \textbf{0.9092 $\pm$ 0.021} & \textbf{0.8961 $\pm$ 0.016} \\
                      &  & Multi  & 0.7615 $\pm$ 0.039 & 0.8809 $\pm$ 0.024 & 0.8846 $\pm$ 0.036 & 0.8813 $\pm$ 0.020 & 0.7601 $\pm$ 0.037 & 0.8760 $\pm$ 0.030 & \underline{0.8797 $\pm$ 0.040}  & 0.8763 $\pm$ 0.026  \\
                      &  & Concat & \underline{0.7724 $\pm$ 0.025} & \textbf{0.8937 $\pm$ 0.018} & 0.8745 $\pm$ 0.019 & \underline{0.8833 $\pm$ 0.012} & \underline{0.7702 $\pm$ 0.025} & \textbf{0.8897 $\pm$ 0.022} & 0.8698 $\pm$ 0.023 & \underline{0.8789 $\pm$ 0.016} \\
                      &  & TF     & 0.7252 $\pm$ 0.055 & 0.8567 $\pm$ 0.033 & \underline{0.8849 $\pm$ 0.044} & 0.8680 $\pm$ 0.026 & 0.7219 $\pm$ 0.053 & 0.8522 $\pm$ 0.033 & 0.8792 $\pm$ 0.050 & 0.8628 $\pm$ 0.030 \\

\bottomrule
\end{tabular}
\end{adjustbox}
\end{table*}

\begin{table*}[b!]
\caption{Cross-corpus Validation Results between D-Vlog and LMVD Datasets on Multimodal Features.}
\label{tab-cross_val_dataset}
\centering
\begin{adjustbox}{width=\textwidth}
\begin{tabular}{c c c c c c c | c c c c}
\toprule
\textbf{Fusion} & \textbf{Train} & \textbf{Test} & \textbf{WAA} & \textbf{WAP} & \textbf{WAR} & \textbf{WAF1} & \textbf{UAA} & \textbf{UAP} & \textbf{UAR} & \textbf{UAF1} \\
\midrule

\multirow{2}{*}{LT} & D-Vlog & LMVD  & 0.6617 $\pm$ 0.083 & \underline{0.8319 $\pm$ 0.047} &  0.9016 $\pm$ 0.078 & 0.8529 $\pm$ 0.019 & 0.6668 $\pm$ 0.071 & 0.8249 $\pm$ 0.054 & 0.8986 $\pm$ 0.079 & 0.8472 $\pm$ 0.024 \\
    & LMVD & D-Vlog  & 0.6367 $\pm$ 0.075 & 0.8131 $\pm$ 0.064 & \underline{0.9040 $\pm$ 0.160} & 0.8188 $\pm$ 0.120 & 0.5816 $\pm$ 0.044 & 0.8427 $\pm$ 0.056 & \underline{0.9166 $\pm$ 0.139} & 0.8456 $\pm$ 0.106 \\

\midrule

\multirow{2}{*}{IT} &  D-Vlog & LMVD  & \underline{0.7170 $\pm$ 0.093} & 0.7997 $\pm$ 0.197 & 0.8529 $\pm$ 0.139 & 0.8183 $\pm$ 0.177 & \underline{0.7167 $\pm$ 0.084} & 0.7981 $\pm$ 0.191 & 0.8514 $\pm$ 0.131 & 0.8161 $\pm$ 0.170 \\
      & LMVD & D-Vlog  & 0.6856 $\pm$ 0.031 & 0.8222 $\pm$ 0.021 & 0.9009 $\pm$ 0.050 & 0.8561 $\pm$ 0.017 & 0.6462 $\pm$ 0.045 & \underline{0.8498 $\pm$ 0.022} & 0.9154 $\pm$ 0.044 & \underline{0.8781 $\pm$ 0.020} \\

\midrule

\multirow{2}{*}{IA} & D-Vlog & LMVD  & \textbf{0.7454 $\pm$ 0.042} & \textbf{0.8784 $\pm$ 0.038} & 0.8736 $\pm$ 0.047 & \textbf{0.8715 $\pm$ 0.020} & \textbf{0.7382 $\pm$ 0.043} & \textbf{0.8756 $\pm$ 0.034} & 0.8661 $\pm$ 0.058 & 0.8660 $\pm$ 0.028 \\ 
                      & LMVD & D-Vlog  & 0.6751 $\pm$ 0.028 & 0.8118 $\pm$ 0.022 & \textbf{0.9172 $\pm$ 0.050} & \underline{0.8562 $\pm$ 0.013} & 0.6265 $\pm$ 0.043 & 0.8414 $\pm$ 0.022 & \textbf{0.9294 $\pm$ 0.042} & \textbf{0.8784 $\pm$ 0.016} \\
                      
\bottomrule
\end{tabular}
\end{adjustbox}
\end{table*}

To thoroughly assess the contribution of different commonly used fusion strategies in MMFformer for depression detection, we performed ablation studies on the D-Vlog and LMVD datasets. The outcomes of the ablation studies are presented in Table \ref{tab-ablation_study}. The ablation experiments focused on combining audio and video modalities using four fusion methods: addition (Add), multiplication (Multi), concatenation (Concat), and tensor fusion (TF) network \cite{zadeh2017tensor}. The Concat achieved the best performance on the D-Vlog dataset, scoring WAA of 0.7652, WAP of 0.8794, WAF1 of 0.8833, and UAF1 of 0.9016. The Add method performed as the second-best, with a high WAR of 0.9112 and  WAF1 of 0.8860, showing its capability in recognizing depressive samples. The Add fusion performed best on the LMVD dataset, with a WAA of 0.7930, WAP of 0.8891, WAR of 0.9120, and WAF1 of 0.8998. It also scored the highest UAF1 of 0.8961. The Concat was the second-best on LMVD, with WAP of 0.8937 and WAF1 of 0.8833, illustrating its consistent performance across datasets. Considering TF fusion, it performed the worst on both datasets, with WAF1 of 0.7357 and 0.8680 on the D-Vlog and LMVD, highlighting its weakness in fusing multi-modal features effectively. In a similar way, the Multi fusion underperformed, achieving WAF1 values of 0.8333 for D-Vlog and 0.8813 for LMVD. These findings revealed that the Concat and Add fusions are more capable at capturing and fusing the complementary information from audio and video modalities. However, our proposed architecture exceeds all these outcomes, where the IT fusion achieves WAA of 0.8108 and WAF1 of 0.9092 on D-Vlog, and the LT method on LMVD records WAA of 0.8071 and WAF1 of 0.9048, demonstrating superior and consistent performance in detecting depression.

\subsection{Cross-Corpus Validation}

To evaluate the generalizability of MMFformer across different datasets, a cross-corpus validation is conducted between D-Vlog and LMVD, focusing on multimodal features. The results of cross-corpus experiments are shown in Table \ref{tab-cross_val_dataset}. Three fusion methods: LT, IT, and IA were tested in two experimental setups: (i) training on D-Vlog and testing on LMVD, and (ii) training on LMVD and testing on D-Vlog. In the first case, when trained on D-Vlog and tested on LMVD, the IA fusion achieved the highest performance, with WAA of 0.7454, WAP of 0.8784, and WAF1 of 0.8715. It also recorded a UAF1 of 0.8660, while the IT method reported a WAA of 0.7170. In the second case, when trained on LMVD and tested on D-Vlog, the IA method again performed well, achieving WAF1 of 0.8562 and high WAR of 0.9172, while the IT fusion recorded WAF1 of 0.8561 and UAF1 of 0.8781. The LT method shows the lowest performance in both scenarios, with WAF1 scores of 0.8529 and 0.8188, respectively. These results indicate that the IA fusion ensured generalizability across multiple datasets, due to its ability to capture and fuse multimodal features effectively. Additionally, the features in D-Vlog appeared more robust for cross-corpus testing, possibly because of its more diverse and realistic content than LMVD. This shows that D-Vlog can significantly enhance research on depression detection in several contexts. 

\section{Conclusion}
\label{Conclusion}
In this paper, a multimodal fusion network called MMFformer is proposed for detecting depression through multiple modalities, including video and audio signals. The video data is processed through a transformer network along with residual connections to extract spatial information. The audio data is exploited by a transformer encoder that helps preserve important information over time, allowing the model to capture significant temporal patterns efficiently. Moreover, the proposed system possessed multimodal capabilities, combining features from multiple modalities through late and intermediate fusion strategies. Experiments on two benchmark datasets reveal that the developed architecture outperforms existing methods in terms of precision of 89.24\% and 90.13\%, recall of 93.80\% and 91.12\%, and F1-Score of 90.92\% and 90.48\% on the D-Vlog and LMVD datasets, respectively. 

We plan to extend this work to evaluate performance using raw data collected from real-life environments to confirm robustness in practical depression detection scenarios. In addition, more data modalities, such as text and physiological data, should be considered to enhance the generalizability of the proposed network. Another essential potential future work is to deploy large language models (LLMs) to enhance better representation in cross-domain depression detection.


\balance
\bibliographystyle{IEEEtran} 
\bibliography{references}    


\end{document}